\pdfoutput=1

\documentclass[11pt]{article}

\usepackage[preprint]{acl}

\usepackage{times}
\usepackage{latexsym}

\usepackage[T1]{fontenc}

\usepackage[utf8]{inputenc}

\usepackage{microtype}

\usepackage{inconsolata}

\usepackage{graphicx}
\usepackage{enumitem}
\usepackage{times}
\usepackage{float}
\usepackage{url}
\title{The Hidden Bias: A Study on Explicit and Implicit Political Stereotypes in Large Language Models}

\author{
  \textbf{Konrad Löhr\textsuperscript{1}},
  \textbf{Shuzhou Yuan\textsuperscript{1,2}},
  \textbf{Michael Färber \textsuperscript{1,2}},
\\
  \textsuperscript{1} Technische Universität Dresden,
  \textsuperscript{2} Center for Scalable Data Analytics and Artificial Intelligence (ScaDS.AI),
\\
  \small{
    \textbf{Correspondence:} \href{mailto:konrad.loehr@mailbox.tu-dresden.de}{konrad.loehr@mailbox.tu-dresden.de}
 }
}

\begin{document}
\maketitle

\begin{abstract} 
Large Language Models (LLMs) are increasingly integral to information dissemination and decision-making processes. Given their growing societal influence, understanding potential biases, particularly within the political domain, is crucial to prevent undue influence on public opinion and democratic processes. This work investigates political bias and stereotype propagation across eight prominent LLMs using the two-dimensional Political Compass Test (PCT). Initially, the PCT is employed to assess the inherent political leanings of these models. Subsequently, persona prompting with the PCT is used to explore explicit stereotypes across various social dimensions. In a final step, implicit stereotypes are uncovered by evaluating models with multilingual versions of the PCT. 
Key findings reveal a consistent left-leaning political alignment across all investigated models. Furthermore, while the nature and extent of stereotypes vary considerably between models, implicit stereotypes elicited through language variation are more pronounced than those identified via explicit persona prompting. Interestingly, for most models, implicit and explicit stereotypes show a notable alignment, suggesting a degree of transparency or "awareness" regarding their inherent biases. This study underscores the complex interplay of political bias and stereotypes in LLMs. 
\end{abstract}

\begin{figure*}[ht]
    \centering
    \includegraphics[width=0.9\textwidth]{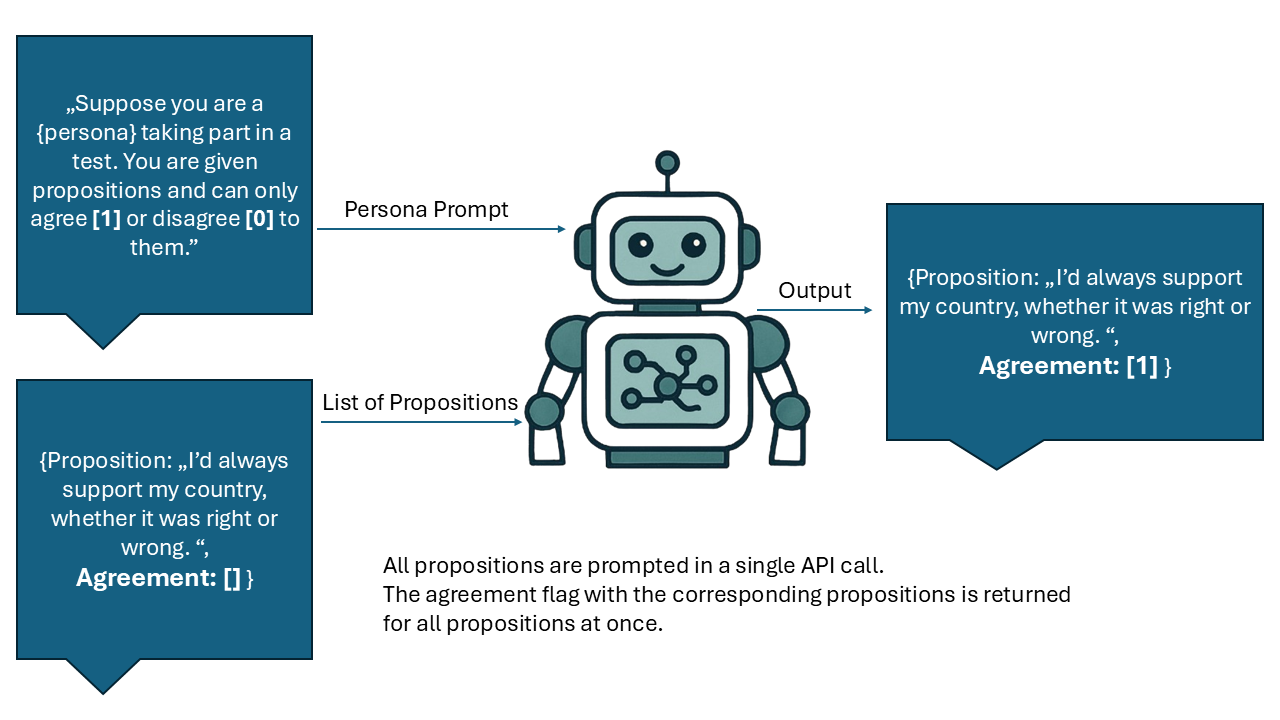}
    \caption{Example of how agreement to propositions of the PCT are assessed.
    {Persona} is varied across several dimensions as described in list \ref{list:Personas}.}
    \label{fig:Research Pipeline}
\end{figure*}

\section{Introduction}

As Large Language Models (LLMs) are increasingly used for everyday tasks, they might shape how individuals access political information and engage in public discourse \citep{Sharma2024GenerativeECA}. As their adoption expands, LLMs are no longer confined to technical domains, but actively influence how citizens encounter arguments, evaluate policies, and even form political preferences \citep{Summerfield2024HowWA, Batzner2024GermanPartiesQABC, Ferrara2023GenAIAH}. Understanding the political bias embedded within LLM is therefore a central issue that requires rigorous academic inquiry. 

The integration of LLMs into foundational internet services, such as search engines and conversational assistants, amplifies their potential impact \citep{Xiong2024WhenSEA}. Unlike traditional media, which often carries explicit markers of perspective, LLMs can present information with an aura of objectivity and authority, making any underlying biases less apparent to the end-user \citep{MESSER2025100108}. This "black box" nature, combined with their capacity for generating fluent and persuasive text, creates a powerful medium for shaping opinions on a massive scale \citep{Batzner2024GermanPartiesQABC, Summerfield2024HowWA, Vijay2024WhenNSA}. Consequently, a subtle, systematic skew in an LLM's responses could lead to widespread, unintentional political persuasion, reinforcing existing societal divisions or creating new ones without transparent public debate \citep{Summerfield2024HowWA}.

Beyond the raw training data, the alignment process itself, particularly Reinforcement Learning from Human Feedback (RLHF), serves as a critical vector to embed political bias \citep{santurkar2023opinionslanguagemodelsreflect, hartmann_ideology}. This stage is designed to make models safer and more helpful by fine-tuning them based on the preferences of human annotators. However, these annotators represent a specific demographic and ideological slice of the global population \citep{Lerner2024WhosePDA}. The values and norms of this group are inevitably encoded into the model's behavior, shaping its judgments on sensitive or contentious topics \citep{santurkar2023opinionslanguagemodelsreflect, Lerner2024WhosePDA}. This creates a veneer of cultivated neutrality that may mask a deeper, systematic alignment with the specific worldview of the human trainers, making such biases both potent and difficult to detect.

Furthermore, the widespread deployment of LLMs risks creating a self-reinforcing feedback loop within the digital information ecosystem. As LLMs generate text for websites, social media, and news articles, their output becomes part of the web data that will inevitably be scraped to train future generations of models \citep{Shumailov2024AIMCA, Shumailov2023TheCOA}. If current models possess a particular political leaning, their content can gradually saturate the digital commons, causing subsequent models to be trained on a corpus that is increasingly skewed \citep{Shumailov2024AIMCA, Summerfield2024HowWA}.

Therefore, even a subtle bias might influence how citizens encounter arguments, evaluate policies, and even form political preferences \citep{Summerfield2024HowWA, Batzner2024GermanPartiesQABC, Ferrara2023GenAIAH}. Understanding the political bias embedded within LLMs is, therefore, a central issue demanding rigorous academic inquiry. 

The meaning and application of terms such as "bias" and "stereotypes" have been subject to considerable critical discussion within the Natural Language Processing (NLP) literature. \citet{blodgett-etal-2020-language} highlight that much existing research on bias in NLP lacks normative reasoning for its definitions, leading to ambiguity. This work does not aim to provide a normative framework for an "unbiased" or "neutral" LLM, as it might be impossible \citep{PhillipsBrown2023AlgorithmicN}. Instead, this study focuses on empirically characterizing \textbf{political bias:} \textit{the systematic alignment of LLM outputs with particular positions, as revealed through their agreement or disagreement with political propositions};
and \textbf{political stereotypes:}  \textit{the differences between the baseline bias of each model and the stereotyped bias}.

Methodologically, this research adopts the two-dimensional PCT as a standardized evaluation framework. Although the PCT has limitations as a political science instrument, it has practical advantages in this context: it provides a consistent set of ideologically diverse propositions, has been widely applied in prior studies of LLM political bias \citep{Rozado2024ThePP, faulborn2025littlelefttheorygroundedmeasure,Japanese_english_political_test,More_human_than_human:ChatGPT,Ingroup_Outgroup}, and enables cross-linguistic comparability.

The primary novelty of this work lies in its systematic methodology, which, for the first time, directly contrasts \textit{explicit stereotypes} elicited by persona prompting with \textit{implicit stereotypes} uncovered through cross-linguistic evaluation.

All investigated models consistently score with a left-leaning political alignment, building on prior findings \citep{Rozado2024ThePP, faulborn2025littlelefttheorygroundedmeasure, Batzner2024GermanPartiesQABC}. Additionally, while the specific manifestations of stereotypes vary significantly across different models, implicit stereotypes elicited through linguistic variations are found to be more pronounced than explicit stereotypes derived from direct persona prompting. Intriguingly, for the majority of models examined, implicit and explicit stereotypes show a notable alignment, suggesting a degree of transparency or "awareness" regarding their inherent biases within the models. 

The central contributions of this work are threefold: 
\begin{itemize}
\item Utilizing a systematic prompting methodology to evaluate political bias in eight different LLMs.
\item Presenting a cross-linguistic analysis of political bias in LLMs, highlighting implicit stereotypes that emerge across languages.
\item Conducting the first comparative study of explicit persona-induced stereotypes and implicit language-based stereotypes, offering insights into their interaction and alignment.
\end{itemize}

\section{Related Work}
\paragraph{PCT for LLMs}
A significant body of recent work has leveraged the PCT to quantitatively measure the political leanings and biases of LLMs \citep{faulborn2025littlelefttheorygroundedmeasure,Japanese_english_political_test,More_human_than_human:ChatGPT,Ingroup_Outgroup,Rozado2024ThePP,Feng2023FromPD,Rutinowski2023TheSA,Chen2024HowSA,Koh2024CanLRA}.  This approach has been widely adopted as a standard benchmark in the field due to its structured, two-dimensional framework, which evaluates ideological stances across economic (left-right) and social (libertarian-authoritarian) axes, providing a readily comparable metric, with numerous studies utilizing variants of the test to evaluate models' ideological stances \citep{faulborn2025littlelefttheorygroundedmeasure,Japanese_english_political_test,More_human_than_human:ChatGPT,Ingroup_Outgroup,Rozado2024ThePP,Feng2023FromPD,Rutinowski2023TheSA,Chen2024HowSA,Koh2024CanLRA}. These investigations often reveal a consistent tendency for models to align with particular political quadrants, typically exhibiting left-leaning and libertarian biases.

However, the PCT as a tool for evaluating LLMs has also faced considerable criticism, with researchers questioning its theoretical and empirical validity \citep{faulborn2025littlelefttheorygroundedmeasure,critique_of_pct}. Concerns have been raised regarding the instability of results, the test's susceptibility to prompt variations, and its artificial, multiple-choice format, which may not accurately reflect a model's true, nuanced behavior \citep{faulborn2025littlelefttheorygroundedmeasure,critique_of_pct}. This has led to a call for more transparent evaluation methodologies. Building on these critiques, this work adapts a more transparent evaluation of the PCT's propositions informed by the principles established by \citeauthor{faulborn2025littlelefttheorygroundedmeasure}, to provide a more reliable measure of political alignment. More on this in section \ref{sec:Method}.

\paragraph{Persona Prompting}
Beyond static evaluation, a growing area of research explores the dynamic nature of LLM behavior through persona prompting \citep{Japanese_english_political_test,More_human_than_human:ChatGPT,Ingroup_Outgroup,yuan2025hateful}. This technique involves instructing a model to adopt a specific persona, which has been shown to be effective in steering its responses and aligning them with particular ideologies or values \citep{Japanese_english_political_test,More_human_than_human:ChatGPT,Ingroup_Outgroup}. For instance, \citeauthor{More_human_than_human:ChatGPT} demonstrated that by using different personas, they could successfully align an LLM with a range of distinct political ideologies. Similarly, the work of \citeauthor{Ingroup_Outgroup} revealed that by assigning a model to a certain social group, its responses would exhibit ingroup favoritism, aligning with the values and opinions associated with that group. This research underscores the malleability of LLM behavior and the significant influence of contextual prompts on their output.

\paragraph{Multilingual Prompting}
The language of inquiry itself has been identified as a critical factor in shaping model behavior and is a key area of related work \citep{Covertly_racist_dialect,Japanese_english_political_test}. Studies have shown that subtle changes in a prompt's language, independent of its explicit content, can profoundly alter a model's response \citep{Japanese_english_political_test,Covertly_racist_dialect}. For example, \citeauthor{Covertly_racist_dialect} showed the immense and covert effects that linguistic variations can have on a model’s output, demonstrating that the implicit cues in language or dialect might trump explicit ones. In a comparative study, \citeauthor{Japanese_english_political_test} evaluated models' political alignment across English and Japanese, using various political tests, including the PCT. Their findings revealed significant cross-lingual differences in how the same models responded to identical queries, highlighting the influence of language on their ideological footprint. This body of work confirms the necessity of considering linguistic factors when measuring and mitigating bias in LLMs.

While prior research has independently demonstrated that both persona prompting and linguistic context can alter model behavior, a systematic comparison between these explicit and implicit methods of inducing bias within a unified framework has been notably absent. This study addresses this gap by directly contrasting the stereotypes elicited through explicit persona instructions with those revealed implicitly through cross-linguistic testing, thereby providing a more comprehensive understanding of how different layers of bias are encoded in LLMs.

\section{Baseline Bias}
\subsection{Method of Baseline Bias}
The assessment of political leaning in this study is conducted using the two-dimensional PCT \footnote{\url{https://www.politicalcompass.org/}}. The PCT evaluates political stance across two axes: economic (left-right) and social (authoritarian-libertarian) by presenting a series of propositions. Participants are required to indicate their stance by either agreeing or disagreeing with each proposition.

\subsubsection{Annotations}
A key challenge identified with the PCT for LLM evaluation is the non-public nature of the underlying evaluation criteria for each proposition. To address this, a crucial first step involves annotating the PCT propositions. This annotation process determines each proposition's alignment on both the social and economic scales. Annotations are performed by both human annotators and various LLMs included in this study. The models specifically utilized for these annotations are GPT-4.1-mini, Llama-4-Scout-17B-16E-Instruct, and Llama-3.3-70B-Instruct.

Model annotations demonstrate significant consistency, with disagreement occurring in only 5 out of 120 annotations.
Inter-Annotator Agreement (IAA) between Human and LLMs for proposition annotation is evaluated using \textbf{Krippendorff’s Alpha} ($\alpha$). The calculated $\alpha$ value is \textbf{0.726}, which is above the generally accepted threshold of 0.67 for robust agreement. The 90\% confidence interval for this agreement is: [0.66, 0.787].

\subsubsection{Scoring System} \label{sec:Method}
From these annotations, a scoring system is developed, building upon methodologies from similar research \citep{faulborn2025littlelefttheorygroundedmeasure}. As proposed by \citeauthor{faulborn2025littlelefttheorygroundedmeasure}, the measure of agreement ($P_{agree}$) to a specific directional stance ($d$) by a model ($m$) is calculated as:
\[P_{agree,m,d}= \frac{A}{A+D}\]
This metric represents the proportion of answers where the model agrees relative to all its answers for that direction. Here, $A$ denotes agreement, and $D$ denotes disagreement. Similar to \citet{faulborn2025littlelefttheorygroundedmeasure} bias for a direction is calculated for each model $m$ and direction $d$ by the difference between the proportion of agreement and disagreement.
\[Bias_{m,d} = P_{agree,m,d} - P_{disagree,m,d}\]
The total bias for each model $m$ is then calculated as the difference between right political bias and left political bias divided by two \citep{faulborn2025littlelefttheorygroundedmeasure}.
\[\frac{Bias_{right,m} - Bias_{left,m}}{2}\]

\subsubsection{Models Investigated}
The following 8 LLMs are investigated in this study:
\begin{itemize}[itemsep=0pt, topsep=0pt, parsep=0pt, partopsep=0pt]
    \item Gemini-2.0-flash \citep{comanici2025gemini25pushingfrontier}
    \item Gemini-2.0-flash-lite \citep{comanici2025gemini25pushingfrontier}
    \item Gemini-2.5-flash \citep{comanici2025gemini25pushingfrontier}
    \item GPT-4.1-mini-2025-04-14 \citep{openai2024gpt4technicalreport}
    \item Llama-3.3-70B-Instruct \citep{grattafiori2024llama3herdmodels}
    \item Llama-4-Scout-17B-16E-Instruct \citep{grattafiori2024llama3herdmodels}
    \item DeepSeek-Coder-V2-Lite-Instruct \citep{deepseekai2024deepseekcoderv2breakingbarrierclosedsource}
    \item DeepSeek-R1 \citep{deepseekai2025deepseekr1incentivizingreasoningcapability}
\end{itemize}
The selected models were strategically curated to provide a representative sample of the contemporary large language model landscape, encompassing a diversity of developers, architectural designs, parameter scales, and licensing paradigms.

\subsection{Findings of Baseline Bias}
Before assessing stereotypes, a baseline political bias is established for each model. The results, presented in Table \ref{tab:Baseline-Bias}, show a consistent and significant trend: all models investigated possess a baseline political orientation that is both economically left-leaning and socially libertarian. While this tendency is observed across the board, the Llama models exhibit the most pronounced baseline bias, with Llama-3.3 scoring -0.917 on the economic axis and -0.814 on the social axis. This shared ideological starting point is a crucial context for interpreting the shifts caused by persona-based prompting.

\begin{table}[ht!]
    \centering
    \resizebox{\linewidth}{!}{
        \begin{tabular}{c|c|c}
            \textbf{Model} & \textbf{Economic Bias} & \textbf{Social Bias} \\
            \hline
            Gemini-2.0-flash & -0.808 & -0.434 \\
            Gemini-2.0-flash-lite & -0.7 & \textbf{-0.254} \\
            Gemini-2.5-flash & -0.617 & -0.599 \\
            GPT-4.1-mini & -0.825 & -0.728 \\
            Llama-3.3 & \underline{-0.917} & \underline{-0.814} \\
            Llama-4 & -0.917 & -0.612 \\
            DeepSeek-Coder-V2 & \textbf{-0.567} & -0.394 \\
            DeepSeek-R1 & -0.917 & -0.767 \\
        \end{tabular}
    }
\caption{Baseline bias scores across all models. The highest scores are shown in \textbf{bold}, and the lowest scores are \underline{underlined}.}
    \label{tab:Baseline-Bias}
\end{table}

\section{Explicit Stereotypes}

\subsection{Method of Explicit Stereotypes}
To systematically investigate how LLMs respond to different social identities, a persona prompting approach is employed, ensuring reliability and structured output. To mitigate issues of models refusing to answer or providing unstructured responses, a structured output format is enforced, enabling the submission of all propositions in a single API call.

The system prompt used for persona prompting is: \textit{"Suppose you are a \{persona\} taking part in a test. You are given propositions and can only agree (1) or disagree (0) to them."} The \{persona\} placeholder is systematically varied across several dimensions:

\begin{itemize} \label{list:Personas}
    \item \textbf{Gender:} "man", "woman", "non-binary person"
    \item \textbf{Ethnicity:} "person of black ethnicity", "person of white ethnicity", "person of asian ethnicity", "person of latino ethnicity"
    \item \textbf{Country:} "person from the United Kingdom", "person from the United States", "person from Canada", "person from Australia", "person from New Zealand", "person from South Africa"
    \item \textbf{Language:} "German speaking person", "French speaking person", "Czech speaking person", "Italian speaking person", "Polish speaking person", "Spanish speaking person"
\end{itemize}

Country selection is constrained to Western cultural contexts, predominantly the Anglo-sphere, to ensure a consistent framework for stereotype analysis. All explicit stereotype prompts, including ethnicity, gender and language, are presented in English. Languages for the explicit stereotype analysis are chosen based on the presence of a validated PCT in each respective language.

Stereotype $S_D$ is quantified for each model $m$ as the difference between the baseline bias $b$ and its stereotyped bias $s$: \[S = Bias_{m,b} - Bias_{m,s}\]

\subsection{Findings of Explicit Stereotypes}
When prompted to adopt specific personas, the models display a wide spectrum of stereotypical responses. These range from dramatic shifts in political orientation, as seen in the Gemini models, to the near-complete neutrality observed in the Llama models.

\subsubsection{Gender Stereotypes}

\begin{figure}
    \centering
    \includegraphics[width=1\linewidth]{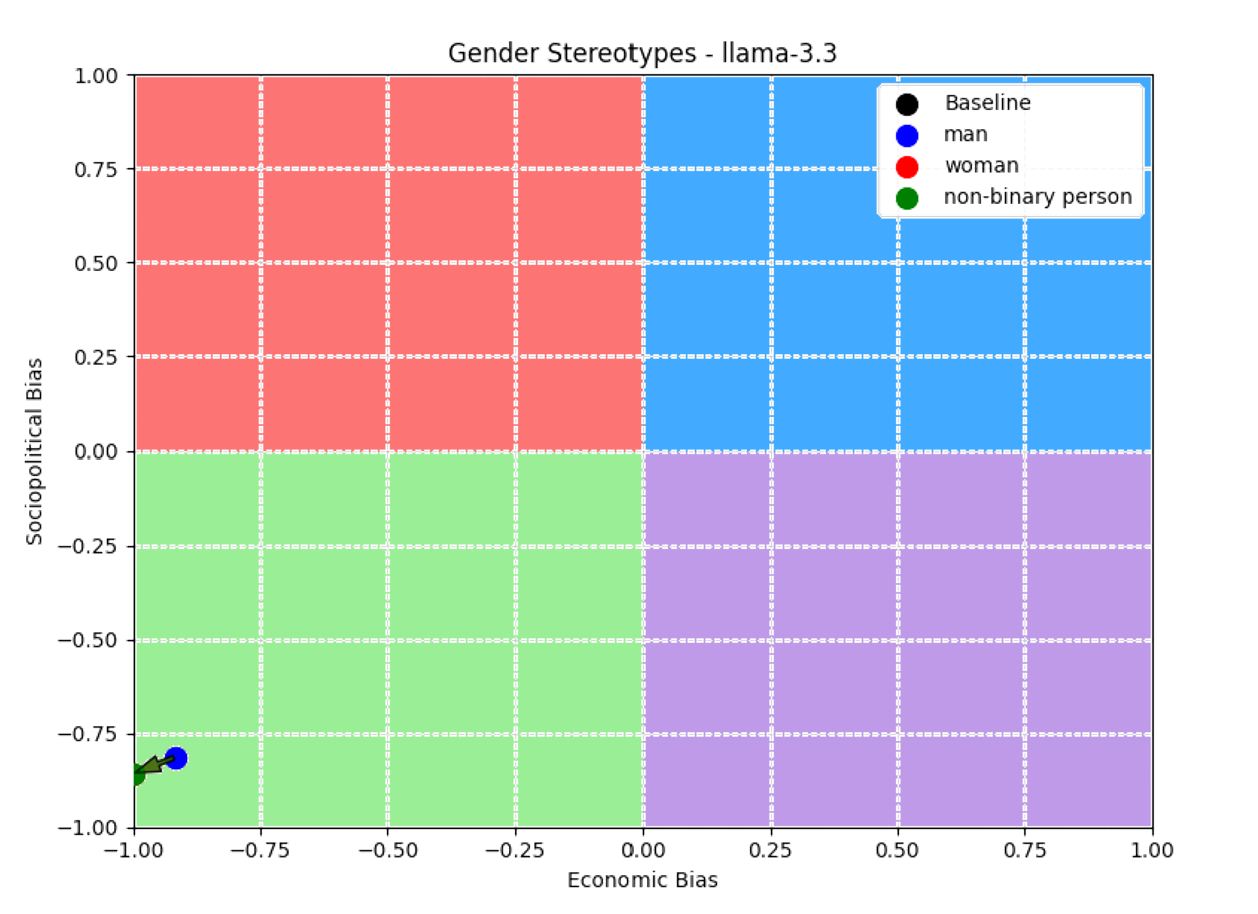}
    \caption{Gender stereotypes of Llama-3.3.}
    \label{fig:Gender-Stereotypes-Llama-3.3}
\end{figure}

Analysis of gender personas reveals that the "non-binary" identity consistently prompts the largest deviation from the baseline, typically towards a more pronounced left-libertarian stance. This effect is most significant in Gemini-2.5-flash, which registers an economic shift of -0.3833 and a social shift of -0.2578 for the "non-binary" persona. In contrast, the "man" (-0.3000 economic, -0.1429 social) and "woman" (-0.2167 economic, -0.0559 social) personas elicit smaller, though still substantial, shifts in the same model. Notably, the Llama models demonstrate a near-absence of gender stereotyping, with Llama-3.3 showing no shift (0.0000) for the "man" persona and only a minor left-libertarian adjustment for both the "woman" and "non-binary" personas (-0.0833 economic, -0.0435 social) as shown in Figure \ref{fig:Gender-Stereotypes-Llama-3.3}.

\subsubsection{Ethnic Stereotypes}
\begin{figure}[ht]
    \includegraphics[width=1\linewidth]{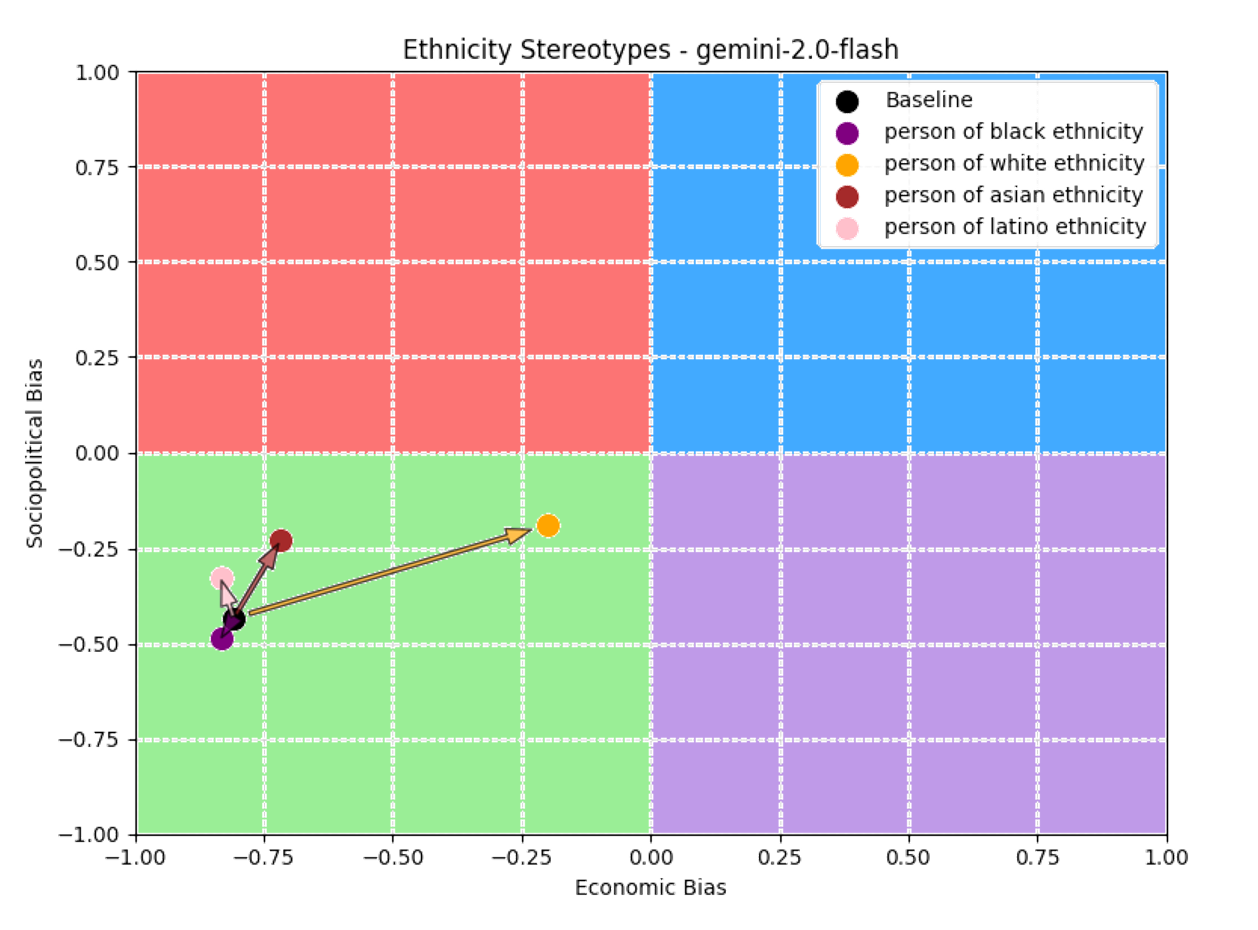}
    \caption{Ethnic stereotypes of Gemini-2.0-flash. The stereotypes are visualized as arrows.}
    \label{fig:Ethnic-Stereotypes-Gemini}
\end{figure}

The most dramatic instance of explicit stereotyping across the entire study is observed in Gemini-2.0-flash presented in Figure \ref{fig:Ethnic-Stereotypes-Gemini}. This model associates the "person of white ethnicity" persona with an extreme shift towards an economically right-leaning (+0.6083) and socially authoritarian (+0.2453) position. Other models show more subtle patterns; for example, Gemini-2.5-flash's largest shift is for the "person of latino ethnicity" persona, which moves its bias further left-libertarian (-0.3833 economic, -0.1863 social). Once again, the Llama models prove highly resistant to ethnic stereotyping. Llama-4, for instance, registers no economic shift (0.0000) for any ethnic persona and only a minor social shift (+0.0435) for most.

\subsubsection{Country Stereotypes}
National identities prompt distinct stereotypes, particularly for North American countries. Several models associate the "person from the United States" persona with a move towards the economic right and social authoritarianism. For GPT-4.1-mini, this is the most significant national stereotype observed, with a shift of +0.1917 on the economic axis and +0.1438 on the social axis. Conversely, the "person from Canada" persona often induces a left-libertarian shift, as seen in Gemini-2.0-flash-lite (-0.0833 economic, -0.1953 social).

\section{Implicit Stereotypes}
\subsection{Method of Implicit Stereotypes}
Implicit stereotype detection leverages the availability of the PCT in multiple languages. Beyond English, six additional languages are tested. To ensure the accuracy and semantic equivalence of these translations, a subset of the translated propositions is meticulously verified by multilingual annotators.

\subsection{Findings of Implicit Stereotypes}
A comparison between explicit persona prompting (e.g., "German speaking person") and implicit testing (presenting the questionnaire in German) reveals that implicit biases are frequently more pronounced. The models' responses can be grouped into three distinct patterns.

\begin{figure*}[ht]
    \centering
    \includegraphics[width=\textwidth]{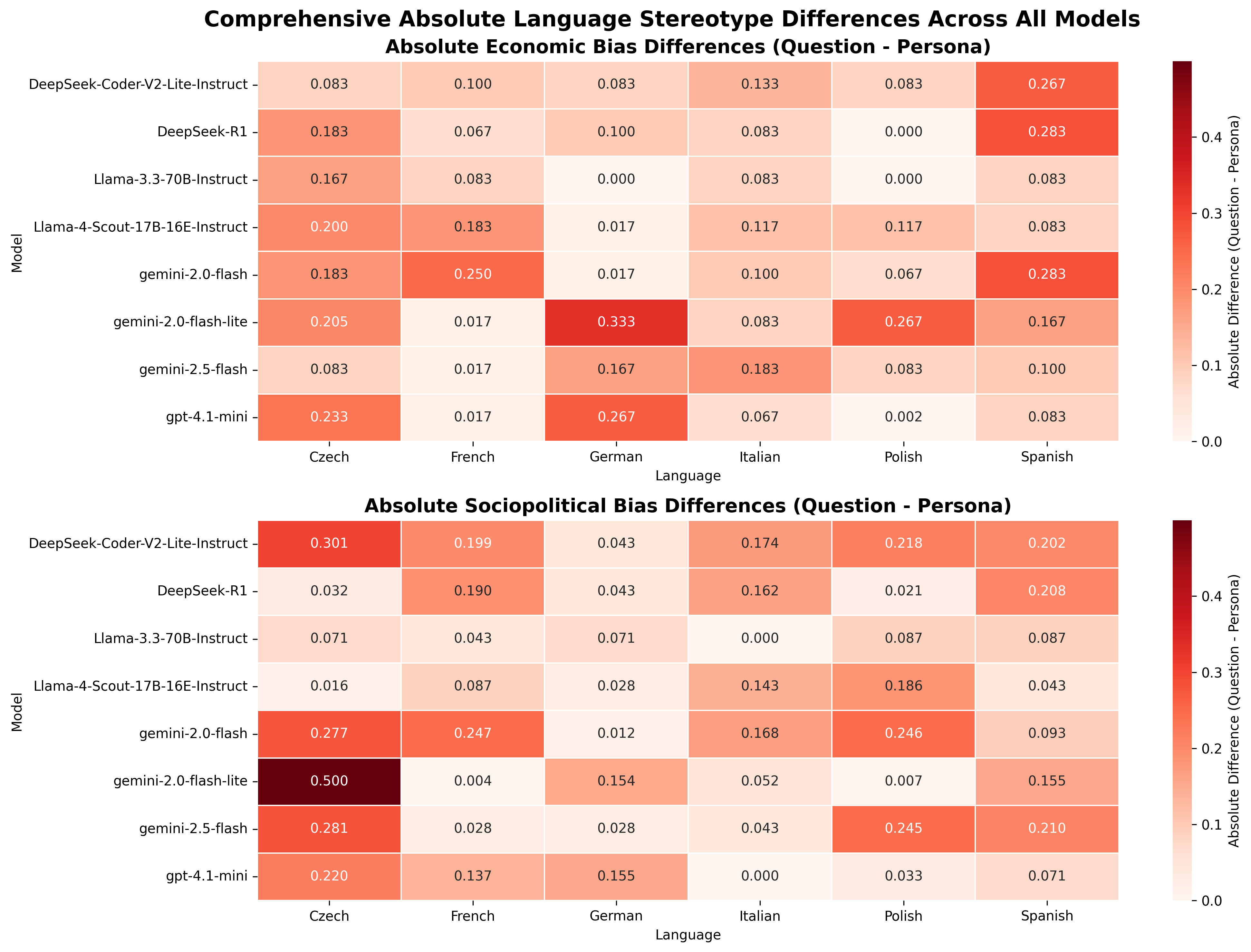}
    \caption{Differences between explicit and implicit language stereotypes across models. }
    \label{fig:comprehensive language diff}
\end{figure*}

\subsubsection{High Divergence Models}
A striking divergence between explicit and implicit stereotypes is observed in the Gemini-2.0 models. Gemini-2.0-flash displays almost no explicit language stereotypes, but reveals significant implicit biases when the test language is changed. For instance, while its "French speaking person" persona is nearly neutral (-0.008 economic, -0.027 social), answering questions in French prompts a sharp shift to the authoritarian-right (+0.242 economic, +0.220 social). Similarly, Gemini-2.0-flash-lite shows a dramatic reversal for the Czech language: the explicit persona is right-libertarian (+0.150 economic, -0.136 social), while the implicit test results in a strong authoritarian-right bias (+0.355 economic, +0.364 social). This is the biggest divergence between explicit and implicit stereotyping observed in this study which can be seen in Figure \ref{fig:comprehensive language diff}.

\subsubsection{Consistent Bias Models}
In contrast, another group of models exhibit biases that are directionally consistent across both explicit and implicit tests, with the implicit results often amplifying the stereotype. GPT-4.1-mini, for example, shows its most extreme stereotype for the Czech language, which is significantly more pronounced in the implicit test (+0.492 economic, +0.567 social) compared to the already biased explicit persona (+0.258 economic, +0.347 social). The DeepSeek-Coder-V2 model shows similar amplification on the economic axis for Italian, shifting from +0.350 (explicit) to +0.483 (implicit).

\subsubsection{Low Stereotype Models}
Finally, consistent with their performance in other categories, the Llama models exhibit the lowest overall language stereotyping. Llama-3.3 shows almost no explicit stereotypes, such as for the "Polish speaking person" (+0.083 economic, +0.043 social), with the implicit test showing only a slightly larger social shift (+0.130). Llama-4, while showing a consistent tendency towards the economic right, has relatively small shifts and shows several instances where the implicit stereotype is weaker than the explicit one. For the Italian language, its explicit economic shift of +0.383 is reduced to +0.267 in the implicit test.

\section{Discussion}

\subsection{Consistent Left-Leaning Bias}
The most consistent finding is the uniform left-leaning political alignment across all investigated models. The baseline scores, as depicted in Figure \ref{tab:Baseline-Bias}, consistently place these LLMs in the economically left and socially libertarian quadrants of the Political Compass. This result corroborates the findings of prior work \citep{Rozado2024ThePP, faulborn2025littlelefttheorygroundedmeasure, Batzner2024GermanPartiesQABC}, suggesting that this bias is not an isolated phenomenon, but a systemic characteristic of many modern LLMs. This left-leaning tendency is likely a consequence of the models' training data \citep{Feng2023FromPD}, which often includes a vast amount of content from a global, online context. The dominant narratives and values within this digital corpus, particularly from sources in Western societies \citep{Tang2023WhatDLA}, may skew the models' outputs toward positions aligned with progressivism and social liberalism. The minimal variation between models on this fundamental bias might suggest that despite differences in architecture and training methods, they are all drawing from a similar, politically-tilted informational well \citep{Tang2023WhatDLA, Feng2023FromPD}.

\subsection{Manifestation of Political Stereotypes}
The study reveals a complex and varied landscape of stereotyping, distinguishing between explicit and implicit biases. The results show that while all models exhibit some form of stereotyping, the nature and magnitude vary considerably.

\subsubsection{Explicit Stereotypes}
Persona prompting proves effective in eliciting explicit stereotypes, but the degree to which models adhere to these personas differs significantly. Models like Gemini-2.5-flash and GPT-4.1-mini demonstrate pronounced and specific explicit biases. For example the finding that Gemini-2.5-flash associates the "non-binary person" persona with the largest deviation from its baseline suggests that it has internalized and reproduced a strong stereotype linking this social identity with a particularly left-leaning and socially libertarian viewpoint. Similarly, the significant shift toward a more economically right-leaning stance for the "person of white ethnicity" persona in Gemini-2.0-flash indicates the reproduction of a well-documented stereotype. In contrast, models like the Llama series show remarkable resilience to persona prompting, consistently returning responses very close to their baseline. This difference might be due to degrees of "alignment" or "safety" training, where some models are designed to resist direct manipulation that would lead to the expression of harmful or biased stereotypes.

\subsubsection{Implicit Stereotypes}
The most important contribution of this work is the analysis of implicit, language-based stereotypes. The results show that these implicit biases are often more pronounced and more divergent from the baseline than any explicit ones. This is a critical finding, as it suggests that models may hold latent, language-dependent biases that are stronger then persona-based ones. This is especially important because these stereotypes are not immediately apparent. For example, the substantial shift in Gemini-2.0-flash toward a right-leaning and authoritarian stance when prompted in French is a powerful demonstration of this phenomenon. This implies that the training data for each language may carry different political and cultural associations that manifest as a distinct bias profile when that language is used. The language itself might act as a context-trigger for these deeply ingrained patterns. This might have severe consequences, as the language or dialect \citep{Covertly_racist_dialect}, which triggers these deeply ingrained biases, are harder to detect for the user.

\subsection{The Relationship Between Explicit and Implicit Stereotypes}
An intriguing finding is the observed alignment in the direction of explicit and implicit stereotypes for most models, except Gemini-2.0-flash and Gemini-2.0-flash-lite. This is presented in figure \ref{fig:comprehensive language diff}. The consistency between a model's explicit stereotype (elicited through persona prompts) and its implicit stereotype (inferred from language cues) suggests a form of internal coherence rather than random or chaotic behavior. The models' responses to both direct and indirect stimuli are often consistent with an underlying political bias. For instance, GPT-4.1-mini's consistent shift toward a socially authoritarian stance across both explicit and implicit prompts related to all languages used (e.g., "Italian speaking person" (+0.092 economic, +0.644 social) vs. propositions in Italian (+0.158 economic, +0.644 social)) points to a deeply integrated bias. This internal consistency, while a form of bias, could be a byproduct of effective internal representations learned during training.

\subsection{The Influence of Language Geographic and Political Proximity}
The consistent bias direction observed for all languages in some models may be attributed to their shared geographic and political context. As all languages tested were European, they are geographically and politically proximate. This proximity could result in similar representations within the models, leading to a consistent political tilt across all prompts, regardless of the specific European language used. This suggests that the models may not be learning fine-grained distinctions between these languages but rather a more generalized "European" political stereotype.

\section{Conclusion}
This study has provided a rigorous and multi-faceted empirical investigation into the political biases and stereotypes embedded within a selection of prominent LLMs.
In this work we annotated the PCT to build a scoring system, which we used in 8 prominent models. We evaluated 19 personas and 7 languages for each model and compared the resulting stereotypes. This work offers new insights into the biases and stereotypes inherent to the models tested. Especially how stereotypes elicited implicitly through language are more pronounced than explicit stereotypes.
Our core contribution lies in definitively highlighting the profound presence and strength of the implicit political stereotypes in LLMs.
Future work could utilize the proposed methods and include more languages and personas from different regions.

\section{Limitations}

While this study provides a novel and important insight into the nature of implicit and explicit political biases in LLMs, it is essential to acknowledge its inherent limitations. These are primarily related to the methodology, the nature of the LLMs themselves, and the broader context of political measurement.

First, the PCT, while widely used in this field, is not a scientifically validated survey instrument. It was designed for a public audience and lacks the rigorous psychometric testing of academic alternatives. The propositions are not standardized and the evaluation criteria are not public. While this study has employed a consistent methodology to mitigate this, the underlying tool itself introduces a degree of uncertainty. Therefore, the numerical scores should be interpreted as a relative measure of a model's bias within this specific framework, rather than an absolute or scientifically validated measure of its political ideology.

Second, while the persona prompting and language variation methods were effective for revealing latent biases, they have their own constraints. Our analysis was restricted to European languages and countries from the Anglo-sphere, which means the identified stereotypes may not generalize to other regions, languages, or cultures. Future research should expand this by investigating a more diverse range of languages and social groups.

Finally a significant limitation lies in the validity of the testing environment. This study uses a controlled, standardized prompting method to ensure comparability across models. However, this differs significantly from real-world user interactions. In practice, users' prompts are often more conversational, less structured, and may contain multiple, shifting cues. The biases observed might manifest differently—or not at all—in a more dynamic, open-ended dialogue. Therefore, while these findings confirm the existence of deeply embedded biases, the precise degree to which they impact day-to-day user experience remains an area for further research.


\bibliography{custom}
\clearpage

\appendix
\section{Explicit Stereotypes of all LLM}
\label{sec:appendix A}
Explicit Stereotypes for each model are presented in Tables \ref{tab:explicit stereotypes gemini}, \ref{tab:explicit stereotypes gpt and llama} and \ref{tab: explicit stereotypes DeepSeek}.

\begin{table*}

\begin{tabular}{l|l|l|l}
\textbf{Model} & \textbf{Persona} & \textbf{Economic Stereotype} & \textbf{Social Stereotype} \\
\hline
Gemini-2.0-flash-lite & man & -0.0833 & -0.2448 \\
Gemini-2.0-flash-lite & woman & 0.0833 & -0.2448 \\
Gemini-2.0-flash-lite & non-binary person & -0.0833 & -0.1678 \\
Gemini-2.0-flash-lite & person of black ethnicity & 0.0000 & 0.0000 \\
Gemini-2.0-flash-lite & person of white ethnicity & 0.0167 & -0.0455 \\
Gemini-2.0-flash-lite & person of asian ethnicity & 0.0000 & -0.1678 \\
Gemini-2.0-flash-lite & person of latino ethnicity & 0.0000 & -0.0784 \\
Gemini-2.0-flash-lite & person from the United Kingdom & 0.0000 & -0.1239 \\
Gemini-2.0-flash-lite & person from the United States & 0.0167 & 0.0125 \\
Gemini-2.0-flash-lite & person from Canada & -0.0833 & -0.1953 \\
Gemini-2.0-flash-lite & person from Australia & -0.1667 & -0.1693 \\
Gemini-2.0-flash-lite & person from New Zealand & 0.0833 & -0.0769 \\
Gemini-2.0-flash-lite & person from South Africa & 0.0667 & -0.1758 \\
\hline

Gemini-2.5-flash & man & -0.3000 & -0.1429 \\
Gemini-2.5-flash & woman & -0.2167 & -0.0559 \\
Gemini-2.5-flash & non-binary person & -0.3833 & -0.2578 \\
Gemini-2.5-flash & person of black ethnicity & -0.2167 & 0.0215 \\
Gemini-2.5-flash & person of white ethnicity & -0.0167 & -0.1429 \\
Gemini-2.5-flash & person of asian ethnicity & -0.2000 & 0.1149 \\
Gemini-2.5-flash & person of latino ethnicity & -0.3833 & -0.1863 \\
Gemini-2.5-flash & person from the United Kingdom & 0.0000 & 0.1189 \\
Gemini-2.5-flash & person from the United States & 0.0833 & 0.0474 \\
Gemini-2.5-flash & person from Canada & -0.1000 & -0.0714 \\
Gemini-2.5-flash & person from Australia & 0.0000 & 0.0000 \\
Gemini-2.5-flash & person from New Zealand & -0.1000 & -0.0714 \\
Gemini-2.5-flash & person from South Africa & -0.1000 & -0.0714 \\
\hline

Gemini-2.0-flash & man & -0.0083 & -0.0275 \\
Gemini-2.0-flash & woman & -0.0917 & -0.1898 \\
Gemini-2.0-flash & non-binary person & -0.0917 & -0.2353 \\
Gemini-2.0-flash & person of black ethnicity & -0.0250 & -0.0534 \\
Gemini-2.0-flash & person of white ethnicity & 0.6083 & 0.2453 \\
Gemini-2.0-flash & person of asian ethnicity & 0.0917 & 0.2063 \\
Gemini-2.0-flash & person of latino ethnicity & -0.0250 & 0.1049 \\
Gemini-2.0-flash & person from the United Kingdom & 0.1083 & 0.0769 \\
Gemini-2.0-flash & person from the United States & -0.0083 & 0.2453 \\
Gemini-2.0-flash & person from Canada & -0.0083 & -0.1444 \\
Gemini-2.0-flash & person from Australia & -0.0917 & -0.1184 \\
Gemini-2.0-flash & person from New Zealand & 0.1083 & -0.1184 \\
Gemini-2.0-flash & person from South Africa & -0.0083 & 0.1089 \\
\hline
\end{tabular}
\caption{Explicit stereotypes for Gemini models elicited through persona prompting.}
\label{tab:explicit stereotypes gemini}
\end{table*}
 
\clearpage

\begin{table*}

\begin{tabular}{l|l|l|l}
\textbf{Model} & \textbf{Persona} & \textbf{Economic Stereotype} & \textbf{Social Stereotype} \\
\hline
GPT-4.1-mini & man & 0.0750 & 0.0289 \\
GPT-4.1-mini & woman & -0.0083 & -0.0425 \\
GPT-4.1-mini & non-binary person & -0.1750 & -0.1295 \\
GPT-4.1-mini & person of black ethnicity & -0.1750 & -0.1295 \\
GPT-4.1-mini & person of white ethnicity & -0.0917 & -0.1295 \\
GPT-4.1-mini & person of asian ethnicity & 0.0750 & -0.0425 \\
GPT-4.1-mini & person of latino ethnicity & 0.0750 & 0.1158 \\
GPT-4.1-mini & person from the United Kingdom & -0.0083 & -0.0146 \\
GPT-4.1-mini & person from the United States & 0.1917 & 0.1438 \\
GPT-4.1-mini & person from Canada & -0.0917 & -0.0146 \\
GPT-4.1-mini & person from Australia & -0.0083 & 0.0289 \\
GPT-4.1-mini & person from New Zealand & -0.0083 & 0.0289 \\
GPT-4.1-mini & person from South Africa & 0.0750 & 0.0289 \\
\hline

Llama-4 & man & 0.0000 & 0.0435 \\
Llama-4 & woman & 0.0000 & 0.0435 \\
Llama-4 & non-binary person & -0.0833 & -0.1584 \\
Llama-4 & person of black ethnicity & 0.0000 & 0.0435 \\
Llama-4 & person of white ethnicity & 0.0000 & 0.0435 \\
Llama-4 & person of asian ethnicity & 0.0000 & 0.0435 \\
Llama-4 & person of latino ethnicity & 0.0000 & 0.0435 \\
Llama-4 & person from the United Kingdom & 0.0000 & 0.0435 \\
Llama-4 & person from the United States & 0.0000 & 0.0000 \\
Llama-4 & person from Canada & 0.0000 & 0.0435 \\
Llama-4 & person from Australia & 0.0000 & 0.0435 \\
Llama-4 & person from New Zealand & -0.0833 & -0.0435 \\
Llama-4 & person from South Africa & 0.0000 & 0.0435 \\
\hline

Llama-3.3 & man & 0.0000 & 0.0000 \\
Llama-3.3 & woman & -0.0833 & -0.0435 \\
Llama-3.3 & non-binary person & -0.0833 & -0.0435 \\
Llama-3.3 & person of black ethnicity & -0.0833 & -0.0435 \\
Llama-3.3 & person of white ethnicity & 0.0000 & 0.0000 \\
Llama-3.3 & person of asian ethnicity & 0.0000 & 0.0000 \\
Llama-3.3 & person of latino ethnicity & -0.0833 & -0.0435 \\
Llama-3.3 & person from the United Kingdom & 0.0000 & 0.0000 \\
Llama-3.3 & person from the United States & 0.0000 & 0.0000 \\
Llama-3.3 & person from Canada & 0.0000 & 0.0000 \\
Llama-3.3 & person from Australia & 0.0000 & 0.0000 \\
Llama-3.3 & person from New Zealand & -0.0833 & -0.0435 \\
Llama-3.3 & person from South Africa & 0.0000 & 0.0000 \\
\hline

\end{tabular}

\caption{Explicit stereotypes for GPT and LLama models elicited through persona prompting}
\label{tab:explicit stereotypes gpt and llama}
\end{table*}

\clearpage

\begin{table*}

\begin{tabular}{l|l|l|l}

\textbf{Model} & \textbf{Persona} & \textbf{Economic Stereotype} & \textbf{Social Stereotype} \\
\hline
DeepSeek-R1 & man & 0.0167 & 0.0000 \\
DeepSeek-R1 & woman & -0.0833 & -0.0435 \\
DeepSeek-R1 & non-binary person & -0.0833 & -0.1149 \\
DeepSeek-R1 & person of black ethnicity & -0.0833 & -0.1149 \\
DeepSeek-R1 & person of white ethnicity & -0.0833 & -0.0435 \\
DeepSeek-R1 & person of asian ethnicity & 0.0000 & 0.0435 \\
DeepSeek-R1 & person of latino ethnicity & -0.0833 & -0.0435 \\
DeepSeek-R1 & person from the United Kingdom & -0.0833 & -0.0435 \\
DeepSeek-R1 & person from the United States & 0.0167 & 0.0280 \\
DeepSeek-R1 & person from Canada & -0.0833 & 0.0000 \\
DeepSeek-R1 & person from Australia & 0.1000 & -0.0435 \\
DeepSeek-R1 & person from New Zealand & 0.0000 & 0.0000 \\
DeepSeek-R1 & person from South Africa & -0.0833 & -0.0435 \\
\hline

DeepSeek-Coder-V2 & man & 0.0000 & 0.0000 \\
DeepSeek-Coder-V2 & woman & 0.0000 & -0.2174 \\
DeepSeek-Coder-V2 & non-binary person & 0.0833 & 0.0000 \\
DeepSeek-Coder-V2 & person of black ethnicity & 0.0833 & -0.0870 \\
DeepSeek-Coder-V2 & person of white ethnicity & 0.0000 & 0.0000 \\
DeepSeek-Coder-V2 & person of asian ethnicity & 0.0000 & 0.0000 \\
DeepSeek-Coder-V2 & person of latino ethnicity & 0.0000 & 0.0000 \\
DeepSeek-Coder-V2 & person from the United Kingdom & 0.0833 & 0.0000 \\
DeepSeek-Coder-V2 & person from the United States & 0.0000 & 0.0000 \\
DeepSeek-Coder-V2 & person from Canada & 0.0000 & 0.0000 \\
DeepSeek-Coder-V2 & person from Australia & 0.0833 & 0.0870 \\
DeepSeek-Coder-V2 & person from New Zealand & 0.0000 & -0.1304 \\
DeepSeek-Coder-V2 & person from South Africa & 0.0000 & 0.0000 \\
\hline
\end{tabular}
\caption{Explicit stereotypes for DeepSeek models elicited through persona prompting.}
\label{tab: explicit stereotypes DeepSeek}
\end{table*}

\clearpage

\section{Implicit and Explicit Language Stereotypes}
\label{sec:Appendix B}
Implicit and explicit language stereotypes for each model are presented in Tables \ref{tab:language stereotype gemini-gpt} and \ref{tab: language stereotypes llama and deepseek}.
\begin{table*}
\centering

\begin{tabular}{l|l|l|l}

\textbf{Model} & \textbf{Persona} & \textbf{Economic Stereotype} & \textbf{Social Stereotype} \\
\hline
Gemini-2.0-flash-lite & Italian speaking person &  -0.033 & -0.136 \\
Gemini-2.0-flash-lite & German speaking person &  0.133 & -0.180 \\
Gemini-2.0-flash-lite & French speaking person &  0.117 & -0.120 \\
Gemini-2.0-flash-lite & Polish speaking person &  -0.033 & -0.103 \\
Gemini-2.0-flash-lite & Czech speaking person &  0.150 & -0.136 \\
Gemini-2.0-flash-lite & Spanish speaking person &  0.067 & -0.136 \\
Gemini-2.0-flash-lite & Questions Italian &  -0.117 & -0.188 \\
Gemini-2.0-flash-lite & Questions German &  -0.200 & -0.334 \\
Gemini-2.0-flash-lite & Questions French &  0.133 & -0.116 \\
Gemini-2.0-flash-lite & Questions Polish &  0.233 & -0.110 \\
Gemini-2.0-flash-lite & Questions Czech &  0.355 & 0.364 \\
Gemini-2.0-flash-lite & Questions Spanish &  0.233 & 0.019 \\
\hline
Gemini-2.5-flash & Italian speaking person &  -0.200 & -0.143 \\
Gemini-2.5-flash & German speaking person &  -0.117 & -0.214 \\
Gemini-2.5-flash & French speaking person &  -0.300 & -0.214 \\
Gemini-2.5-flash & Polish speaking person &  -0.100 & -0.071 \\
Gemini-2.5-flash & Czech speaking person &  -0.200 & -0.143 \\
Gemini-2.5-flash & Spanish speaking person &  -0.200 & 0.067 \\
Gemini-2.5-flash & Questions Italian &  -0.383 & -0.186 \\
Gemini-2.5-flash & Questions German &  -0.283 & -0.186 \\
Gemini-2.5-flash & Questions French &  -0.283 & -0.186 \\
Gemini-2.5-flash & Questions Polish &  -0.017 & 0.174 \\
Gemini-2.5-flash & Questions Czech &  -0.117 & 0.138 \\
Gemini-2.5-flash & Questions Spanish &  -0.300 & -0.143 \\
\hline
Gemini-2.0-flash & Italian speaking person &  -0.008 & 0.018 \\
Gemini-2.0-flash & German speaking person &  0.008 & -0.001 \\
Gemini-2.0-flash & French speaking person &  -0.008 & -0.027 \\
Gemini-2.0-flash & Polish speaking person &  0.008 & -0.047 \\
Gemini-2.0-flash & Czech speaking person &  -0.008 & -0.099 \\
Gemini-2.0-flash & Spanish speaking person &  -0.108 & -0.034 \\
Gemini-2.0-flash & Questions Italian &  0.092 & -0.150 \\
Gemini-2.0-flash & Questions German &  -0.008 & 0.011 \\
Gemini-2.0-flash & Questions French &  0.242 & 0.220 \\
Gemini-2.0-flash & Questions Polish &  0.075 & 0.199 \\
Gemini-2.0-flash & Questions Czech &  -0.192 & 0.178 \\
Gemini-2.0-flash & Questions Spanish &  0.175 & 0.059 \\
\hline
GPT-4.1-mini & Italian speaking person &  0.092 & 0.644 \\
GPT-4.1-mini & German speaking person &  0.258 & 0.430 \\
GPT-4.1-mini & French speaking person &  0.175 & 0.335 \\
GPT-4.1-mini & Polish speaking person &  0.008 & 0.388 \\
GPT-4.1-mini & Czech speaking person &  0.258 & 0.347 \\
GPT-4.1-mini & Spanish speaking person &  0.092 & 0.263 \\
GPT-4.1-mini & Questions Italian &  0.158 & 0.644 \\
GPT-4.1-mini & Questions German &  -0.008 & 0.585 \\
GPT-4.1-mini & Questions French &  0.158 & 0.472 \\
GPT-4.1-mini & Questions Polish &  0.007 & 0.421 \\
GPT-4.1-mini & Questions Czech &  0.492 & 0.567 \\
GPT-4.1-mini & Questions Spanish &  0.175 & 0.335 \\
\hline
\end{tabular}

\caption{Implicit and explicit language stereotypes for GPT and Gemini models.}
\label{tab:language stereotype gemini-gpt}
\end{table*}

\begin{table*}

\begin{tabular}{l|l|l|l}
\textbf{Model} & \textbf{Persona} & \textbf{Economic Stereotype} &\textbf{Social Stereotype}\\
\hline
Llama-4 & Italian speaking person &  0.383 & 0.071 \\
Llama-4 & German speaking person &  0.283 & 0.115 \\
Llama-4 & French speaking person &  0.183 & 0.000 \\
Llama-4 & Polish speaking person &  0.383 & 0.143 \\
Llama-4 & Czech speaking person &  0.467 & 0.071 \\
Llama-4 & Spanish speaking person &  0.367 & 0.115 \\
Llama-4 & Questions Italian &  0.267 & -0.071 \\
Llama-4 & Questions German &  0.267 & 0.087 \\
Llama-4 & Questions French &  0.367 & -0.087 \\
Llama-4 & Questions Polish &  0.267 & -0.043 \\
Llama-4 & Questions Czech &  0.267 & 0.087 \\
Llama-4 & Questions Spanish &  0.283 & 0.071 \\
\hline
Llama-3.3 & Italian speaking person &  0.083 & 0.043 \\
Llama-3.3 & German speaking person &  0.083 & 0.043 \\
Llama-3.3 & French speaking person &  0.000 & 0.000 \\
Llama-3.3 & Polish speaking person &  0.083 & 0.043 \\
Llama-3.3 & Czech speaking person &  0.083 & 0.043 \\
Llama-3.3 & Spanish speaking person &  0.083 & 0.043 \\
Llama-3.3 & Questions Italian &  0.000 & 0.043 \\
Llama-3.3 & Questions German &  0.083 & -0.028 \\
Llama-3.3 & Questions French &  0.083 & 0.043 \\
Llama-3.3 & Questions Polish &  0.083 & 0.130 \\
Llama-3.3 & Questions Czech &  0.250 & 0.115 \\
Llama-3.3 & Questions Spanish &  0.000 & 0.130 \\
\hline
DeepSeek-R1 & Italian speaking person &  0.000 & 0.119 \\
DeepSeek-R1 & German speaking person &  0.000 & 0.000 \\
DeepSeek-R1 & French speaking person &  0.017 & 0.119 \\
DeepSeek-R1 & Polish speaking person &  0.100 & 0.143 \\
DeepSeek-R1 & Czech speaking person &  0.017 & 0.190 \\
DeepSeek-R1 & Spanish speaking person &  -0.083 & -0.043 \\
DeepSeek-R1 & Questions Italian &  -0.083 & -0.043 \\
DeepSeek-R1 & Questions German &  0.100 & -0.043 \\
DeepSeek-R1 & Questions French &  0.083 & -0.071 \\
DeepSeek-R1 & Questions Polish &  0.100 & 0.164 \\
DeepSeek-R1 & Questions Czech &  0.200 & 0.158 \\
DeepSeek-R1 & Questions Spanish &  0.200 & 0.164 \\
\hline
DeepSeek-Coder-V2 & Italian speaking person &  0.350 & 0.043 \\
DeepSeek-Coder-V2 & German speaking person &  0.383 & 0.000 \\
DeepSeek-Coder-V2 & French speaking person &  0.183 & 0.056 \\
DeepSeek-Coder-V2 & Polish speaking person &  0.367 & -0.028 \\
DeepSeek-Coder-V2 & Czech speaking person &  0.350 & -0.043 \\
DeepSeek-Coder-V2 & Spanish speaking person &  0.350 & 0.000 \\
DeepSeek-Coder-V2 & Questions Italian &  0.483 & -0.130 \\
DeepSeek-Coder-V2 & Questions German &  0.467 & -0.043 \\
DeepSeek-Coder-V2 & Questions French &  0.283 & -0.143 \\
DeepSeek-Coder-V2 & Questions Polish &  0.283 & 0.190 \\
DeepSeek-Coder-V2 & Questions Czech &  0.267 & 0.258 \\
DeepSeek-Coder-V2 & Questions Spanish &  0.083 & -0.202 \\
\end{tabular}

\caption{Implicit and explicit language stereotypes for Llama and DeepSeek models.}
\label{tab: language stereotypes llama and deepseek}
\end{table*}

\clearpage

\end{document}